\documentclass[sigconf, screen]{acmart}

\AtBeginDocument{%
  \providecommand\BibTeX{{%
    \normalfont B\kern-0.5em{\scshape i\kern-0.25em b}\kern-0.8em\TeX}}}

\copyrightyear{2021}
\acmYear{2021}
\setcopyright{acmlicensed}\acmConference[FDG'21]{The 16th International Conference on the Foundations of Digital Games (FDG) 2021}{August 3--6, 2021}{Montreal, QC, Canada}
\acmBooktitle{The 16th International Conference on the Foundations of Digital Games (FDG) 2021 (FDG'21), August 3--6, 2021, Montreal, QC, Canada}
\acmPrice{15.00}
\acmDOI{10.1145/3472538.3472602}
\acmISBN{978-1-4503-8422-3/21/08}

\newcommand{\rpm}{\raisebox{.2ex}{$\scriptstyle\pm$}}

\begin{document}

\title{Generating Lode Runner Levels by Learning Player Paths with LSTMs}


\author{Kynan Sorochan}
\affiliation{%
  \institution{University of Alberta}
  \city{Edmonton}
  \country{Canada}}
\email{ksorocha@ualberta.ca}

\author{Jerry Chen}
\affiliation{%
  \institution{University of Alberta}
  \city{Edmonton}
  \country{Canada}}
\email{jerry3@ualberta.ca}

\author{Yakun Yu}
\affiliation{%
  \institution{University of Alberta}
  \city{Edmonton}
  \country{Canada}}
\email{yakun2@ualberta.ca}

\author{Matthew Guzdial}
\affiliation{%
 \institution{University of Alberta}
  \city{Edmonton}
  \country{Canada}}
\email{guzdial@ualberta.ca}

\renewcommand{\shortauthors}{Sorochan et al.}

\begin{abstract}
    Machine learning has been a popular tool in many different fields, including procedural content generation. 
    However, procedural content generation via machine learning (PCGML) approaches can struggle with controllability and coherence.
    In this paper, we attempt to address these problems by learning to generate human-like paths, and then generating levels based on these paths. 
    We extract player path data from gameplay video, train an LSTM to generate new paths based on this data, and then generate game levels based on this path data. 
    We demonstrate that our approach leads to more coherent levels for the game Lode Runner in comparison to an existing PCGML approach.
\end{abstract}





\maketitle

\section{Introduction} 

%%Par 1. Is about situating the area that this research is focused on, and identifying a problem that exists in that area. You want to start broad then get more specific. 
%%You might want to start with ``PCGML'' and then move to talking specifically about level generation.
%%What do we get if we solve this?
%%Problem: playable ML levels that aren't platformers

%%Par 2. Explain to the reader why the problem you identified in par 1 is not yet solved. Lots of ways to make this argument, but the important thing is to convince the reader this is challenging and/or that it represents novelty. 

%%Par 3. Okay, here's what we actually did, and make clear how this work addresses or gets us closer to solving that problem (even though it is very unlikely to have solved it solved it completely).

%%Par 4. (OPTIONAL) Walk the reader through the rest of the paper in terms of its structure, and clearly state the contributions (the specific new and valuable ideas we're putting forward) of this paper. 

Procedural content generation via machine learning (PCGML) is the study and application of machine learning to procedurally generating content, particularly for games \cite{summerville2018procedural}.
While PCGML has had enjoyed considerable popularity recently, a number of open problems exist. 
Particularly in comparison to traditional, non-ML PCG, PCGML approaches struggle with controllability and coherence. 

We define controllability as the ability for a user to impact particular attributes of the generated content. 
In traditional PCG, since the system is authored by a human, there are a number of strategies to allow a user to impact the output or enforce particular constraints \cite{smith2012case,liapis2016mixed,horswill2019imaginarium}.
There are efforts to make PCGML approaches controllable, but this is still an under-explored problem \cite{mott2019controllable,sarkar2020controllable,cheng2020automatic,chen2020image}. 
In particular, we identify a lack of focus on approaches that allow users to specify high-level, intuitive constraints as the input to an ML generator that then outputs game content that matches those constraints. 

By coherence we indicate the problem of game content demonstrating global coherence, global structure that fits human understanding of that game content. 
Global structure includes a wide range of constraints, and is dependent on the particular game content in question.
For example, playability in game levels, the ability to complete said level, is an example of global structure that we expect from human-authored game levels. 
However, a level being playable is not the only element of global structure that we expect. 
A completely flat platformer game level would be playable, but would violate other elements of global structure. 
Modeling global structure is a common problem in machine learning generally \cite{moon2019unified}.
In PCGML, there have been attempts to model global structure, with the most common approach being to model the player's path through a game level \cite{summerville2016super,sarkar2020exploring}.
However, this area is also under-explored. 

In this paper, we investigate a novel PCGML approach that attempts to address these two open problems: controllability and coherence. 
Specifically, we introduce an approach to generate Lode Runner levels based on specified paths. Our focus on coherence will be in reference to Lode Runner Levels.
We model these paths to ensure that they are human-like with a Long Short-Term Memory Recurrent Neural Network (LSTM RNN or LSTM). 
For training data we extract real human paths on existing Lode Runner levels from gameplay video. 
We then use the LSTM to generate novel human-like paths and employ a Markov Chain to generate novel levels based on these generated paths. 
We employ a Markov Chain for this initial investigation as it represents a simple ML model that typically struggles to capture global structure. 
Therefore, it's an ideal choice to investigate whether this approach improves level coherence.
Our approach is inherently controllable as we can input arbitrary paths, though we focus on our generated, human-like paths in this paper. We acknowledge that we won't be directly evaluating the controllability of this approach, but still contend that it is controllable.

In this paper, we first introduce related prior work. 
We then overview our generator, from data extraction to the novel generation of Lode Runner levels. 
We compare the performance of our generator to an existing Markov Chain generator without path data to evaluate whether we demonstrate improved coherence \cite{snodgrass2015hierarchical}. 
We then present a secondary evaluation of our approach in comparison to the original Lode Runner levels. 
We end with a discussion of our limitations and future work.

\section{Related Works}

%Pure novelty argument section, this is a very defensive section.

%Each subsection/paragraph should introduce an area of related work (so for example Loderunner level generation, but you might also have sections on music generation, or ML tools, etc.). Then you'd talk about the general strategies or types of work in this area (without getting into detail, but still citing them). Then identify the 1-3 most similar examples of prior work, and describe them in 1-2 sentences each (only describe the aspects related to your work). Then end off by stating how your work differs still from even this most similar prior work.

In this section we overview work in terms of prior PCGML approaches to generate Loderunner levels, controllability in PCGML, and coherence via paths in PCGML.

Many PCGML approaches have been applied to level generation for Lode Runner in recent years. 
Thakkar et al. proposed the use of a variational autoencoder to generate new levels for Lode Runner based on a binary encoding of each character in the original levels. 
They attempted to improve the playability, one aspect of global structure, by searching the learned, latent space with an evolutionary algorithm \cite{thakkar2019autoencoder}. 
We also attempt to increase playability, but based on altering the input to the generation pipeline instead of including search-based PCG within the pipeline.
Snodgrass and Ontan{\'o}n made use of Markov models to generate content for many games including Lode Runner \cite{snodgrass2016learning}, we include this approach as a baseline as we also make use of a Markov chain for our generator.

Markov chains have been a common method for PCGML since its inception \cite{snodgrass2013generating}.
Much of this prior work has focused on \emph{Super Mario Bros.} level generation, a common area of PCGML research \cite{snodgrass2013generating,snodgrass2014hierarchical,Summerville2015MCMCTSP4,snodgrass2015hierarchical}.
We also employ Markov chains, as they typically struggle with global coherence in comparison to other methods \cite{guzdial2018co}.

Many approaches have been made to attempt to improve the coherence of PCGML output \cite{thakkar2019autoencoder}.
Of particular interest to us are approaches that attempt to do this by invoking some representation of the player path \cite{sarkar2020exploring}. 
Summerville et al. trained an LSTM on Super Mario Bros. levels that included representations of potential player paths \cite{summerville2016super}. 
Follow up work by Summerville et al. extracted player paths from gameplay video and found that these led to significantly different output levels when they were used to train LSTMs \cite{summerville2016learning}.
We use a similar method to extract player paths.
The major difference between Summerville et al.'s approach and ours is that they alter the level representation to include the path information. 
Instead, we generate novel paths and use these as input for a PCGML level generator. 
The Summerville et al. generator could not take a specified path as input without modification. 
This and much of the prior work mentioned above is based on the representations from the Video Game Level Corpus (VGLC) \cite{VGLC}, which we draw on for our level representation.

\section{System Overview}

%Section that describes your approach in sufficient detail that someone could reimplement it based only on this section
%A nice big overview figure here. 

%Argument: anytime you make a claim or discuss a particular detail, you have to back it up

We aim to show the benefits of human-like paths to improve global coherence in PCGML generators.
Our approach can be divided into several steps: (1) extracting the paths from gameplay videos, (2) training an LSTM on the path data, (3) training our Markov chain on the original Lode Runner levels, and (4) generating a new level from a generated path.

\subsection{Data extraction}

Our goal for this first step is to extract human paths for solving the original Lode Runner levels, which we extract from gameplay video.
We focus on human paths instead of paths generated by an automated level playing agent as in prior work \cite{summerville2016super}.
We made this choice as prior work found that human paths led to different output levels than automated paths \cite{summerville2016super}.
Based on this prior work, we make the assumption that automated paths would lead to levels that were more sparse, but more human-like paths will lead to paths closer to the original game. 
We leave a verification of this assumption for future work. 

We download a series of videos from YouTube, based on similar approaches in prior work \cite{guzdial2016toward,summerville2016super},
Once we have the videos, we extract the frames and track the location of the player in each frame for each level, based on the same approach used in the above prior work.
We tag each location with the type of movement using OpenCV and pattern matching \cite{bradski2000opencv}. 
We do this by hand tagging a series of images or sprites representing the different actions of the player in Lode Runner.
In each frame, we identify the player's action or type of movement based on the image with the highest probability.
We map the player's location in the frame to a 32 by 22 grid, which is the size of the VGLC dataset for each level.
This gives us a sequence of 32 by 22 grids equal in length to the amount of time the player played a given level.
After obtaining this sequence we can identify one out of five possible actions for each grid (moving left as 'l', moving right as 'r', climbing up 'u', climbing down 'c', and falling down as 'f') based on the location change of the player between pairs of grids. 
This allows us to store a player's path for each level as a one-dimensional sequence. 

\subsection{Path generation}

We could have simply reused the extracted player paths as the input for a Lode Runner level generation process. 
However, this would have limited the number of levels our system could generate.
As such, we need some way to generate new human-like paths. 

Our extracted human paths vary significantly both in terms of length and patterns.\cite{github} 
To address this, we first split the sequences of actions into small chunks of a fixed size (50). 
We represent each action as a one-hot encoding of length five for the five actions. 
Every action becomes a vector of length five, with every action type represented as an index in that vector.
An action is represented by a 1 at its index if it occurred at this position in the sequence and a 0 otherwise. 
Thus our data becomes a series of 50x5 matrices. 

We employ a Long Short-Term Memory Recurrent Neural Network (LSTM) for our path modeling, as they have been demonstrated to work well on PCGML tasks with sequence-like data \cite{summerville2016learning}.
An LSTM is designed to better learn long-term dependencies than standard recurrent neural networks.
This is important for our use case as the human paths tended to have many repeated actions in a row, and we didn't want our model to learn to just repeat the same action endlessly. 

We employ an LSTM-based Seq2Seq model that is built to generate new paths. 
Our input is one 50-length path sequence and the expected output is the next 50-length path sequence. 
The model is composed of 2 layers with 512 LSTM cells each. 
We employ dropout and gradient clipping to prevent overfitting and gradient explosion, respectively. 
The final layer is a fully connected layer of length 50 with softmax activation, to better represent the probability distribution of a single character at each time step. 
We trained our model for 60 epochs with a 0.001 learning rate and the adam optimizer\cite{kingma2017adam}. 
We probabilistically sample each output action based on treating the softmax activation as a probability distribution.
We note that despite using 50 inputs and 50 outputs this model can be used to create paths of arbitrary lengths by inputting empty (all 0s) actions initially, and then continually generating based on previously generated outputs.  

\subsection{Level Structure Learning}

The path information we generate only partially defines a level.
The same level path could be associated with a large, but not infinite number of levels.
This is particularly true in Lode Runner, where the player can directly ``make'' paths through their own actions (e.g. digging holes). 
This means that we still need more information about what kind of tiles can be associated with what action, and how to fill out the rest of the level ``away'' from the player path. 
Given our interest in demonstrating the impact of player paths in improving global coherency, we employ a multi-dimensional Markov chain \cite{snodgrass2014experiments}. 
In our multi-dimensional Markov chain each tile value depends on the tiles to its left and directly below, along with the player action at the current position, to the right, and above.
We note that prior examples of platformer Markov chains have made use of nodes with 3 dependencies: the left, below, and to the left and below. 
We experimented with this model, but found that our 2-tile dependency sufficiently modeled level structure and led to increased diversity in the output.
We train our multi-dimensional Markov chain on the Lode Runner levels from the VGLC \cite{VGLC}, with the added information from our extracted paths in the grid-based representation discussed above.

We also record a series of statistics in terms of the number of enemies and gold pieces in the training levels. 
For each original level, we found the ratio of the numbers of enemies and gold pieces to the associated player path lengths. 
We represent these two distributions as two Gaussian distributions.
This allows us to sample from these two distributions and derive an overall number of desired gold pieces and enemies for a new level.

\subsection{Complete Level Generation}

Our level generation process begins by generating a new path.
From this path we can determine the minimum size of a level necessary to contain this path.
We instantiate this level as an empty grid of the minimum size.
We label every visited tile in this level with the action that the player path indicates, and the remaining tiles with a special token that indicates no actions.
We automatically constrain some tile values based on the action type, based on the existing keys in the Markov Chain.
For example, if the action taken at a tile is to move up, then we know that this tile must contain a ladder. 
Or if the action at a tile is to move to the left, then this tile can be empty, have a brick (as the player may break it from above and fall to it, then move left), or a rope. 
This gives us some high-level constraints on the tiles and a basic initial structure to work with.

We apply our Markov chain to fill the level out with the final tile types from the partially specified state.
We start from the bottom left corner and move along each row from left to right until we've reached the top of the screen. 
If a tile has been specified we do not need to generate a new tile at this location, and we just move on. 
If the tile has been partially specified then we remove the tile possibilities that have already been ruled out, and then probabilistically sample from the remaining options based on the learned probability distribution in the Markov chain. 
If the tile has not been specified at all, then we simply probabilistically sample from the Markov chain as normal \cite{snodgrass2014experiments}.
In the event that a key does not exist, we remove all dependencies except for the left and below dependency and then sample from this simplified distribution.
We made this choice as this removes all path requirements and only considers the structure of the level.

The final step of our generation process is to place all enemy and gold tiles. 
We randomly sample from both of the Gaussian distributions we described above.
We multiply the sampled values by the generated path length to get the final numbers of enemies and gold pieces.
We then randomly place these elements along the player's path, as they are elements that the player would have to avoid or seek out, respectively.
We made this choice as the Lode Runner levels in the VGLC only place enemies and gold pieces in place of empty tiles, but both entities can move. 
Further, the Markov chain struggled to learn to place these elements effectively given how rare they are in comparison to the other tile types.

\section{Quantitative Evaluation}

%Outline and argue as to why the evaluation that you used is appropriate
%Remind the reader what we were trying to accomplish

%Give the basic task of the evaluation, generating Loderunner levels and comparing in terms of playability
%Give the baselines that we used and why they're appropriate (also indicate why we didn't use baselines that might seem "obvious"). 

%Then give those metrics and explain why we're including these metrics

%Initial paragraph where we re-introduce the overall focus of this paper, so for example saying that this work centres on a model that takes as input a player path and outputs a level that incorporates that path. 
%In this section we evaluate this model in terms of [comparing its output levels to a generator without player path input.]
%[Introduce our baseline approach AKA Sam Snodgrass' work.]
%And why is this evaluation approach reasonable.

\begin{figure}[tbh]
    \centering
    \includegraphics[width=3in]{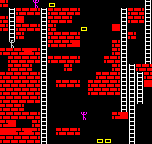}
    \caption{A good generated level.}
    \label{fig:good1}
\end{figure}

We extracted 66 player paths for 66 game levels from a 4.5-hour long gameplay video. 
After training on these sequences, we were able to generate an arbitrary number of levels. 
Figure \ref{fig:good1} gives an example of a generated level based on our approach.

This research seeks to determine whether adding a player path as input to a PCGML generator improves global coherence. 
We employ Markov chains, a PCGML approach that tends to struggle with global coherence in order to better understand the impact of player paths \cite{guzdial2018co}.
As such, the natural choice for an initial evaluation is investigating how the inclusion of a generated path as input changes the generated levels in comparison to a Markov chain approach without player path information.
We therefore compare against the work of Snodgrass and Ontan{\'o}n, who employed a Markov chain without path information to generate Lode Runner levels \cite{snodgrass2016learning}.

%we generated 38 player paths in total, which resulted in 38 levels out of which 31 levels are playable. Figure \ref{fig:good1} shows a playable results by our method.
We employed an A* pathfinding agent to test how a player might play the two different sets of Lode Runner levels. \cite{github} 
Ideally, we might have used a human subject study to compare the two types of generated levels. 
However, we lacked the time and resources for a human subject study, and so use this method for an initial comparison.
The pathfinder starts from the player tile present in both sets of generated levels and attempts to pathfind to each gold piece in turn.
Lode Runner is a complex game, which allows players to momentarily trap enemies, as such, we ignore enemies while pathfinding.
This decision was also motivated by the fact that we lacked a simulator to fully simulate enemy movement in the game, meaning that the enemies would just be treated as impassable obstacles otherwise, which would not be appropriate.
We note that prior work did not make this assumption, and so they reported much lower percentages of playable outputs \cite{snodgrass2016learning}. 
However, since this assumption is made for both types of levels it's still helpful for comparison purposes.
The pathfinder tracks the number of nodes explored on the way to each gold as well as each gold that it was able to reach.  
It reports the total number of each of those two metrics which we use to compare between the two types of levels.  

Ideally, an A* agent should be able to reach each gold piece from the starting location in a playable Lode Runner level.
We had the A* agent report the number of nodes explored on the way to each gold piece as a measure of coherence. 
All of the existing Lode Runner levels have clear paths to access each gold piece, essentially acting as puzzles for the player to solve. 
Therefore, we take a low number of nodes explore as an indirect measure of global coherence.

%We need to be able to automatically test levels in terms of player paths. 
%[Talk about your A* agent and how it works, and why the decisions you made for it are appropriate.]

%[We use the A* agent to collect the following metrics, as approximations for playability and player experience:]
%Using this A-star agent, we collected the following metrics that allowed us to see if our levels could outperform the Snodgrass levels.
%For each metric, introduce it and how you calculate it. And back up why this metric is useful or what it will tell us about the two generators. 
We employ the following metrics in this comparative evaluation, reporting the average and standard deviation of each metric across the generated levels:
\begin{itemize}
    \item \textit{Gold Total Per Level} - This is the average total number of gold pieces the A* agent needed to find. This is used in determining the percentage of gold collected in a level, as well as the potential difficulty and length of a level.  The more gold pieces the more potential for difficulty depending on how they are distributed throughout the level. It also coincides with how long a level may take to play. The more gold, the longer it could take for a player to collect each piece.  
    \item \textit{Percentage Collected Per Level} - This metric gives how many of the gold pieces could be collected in a level.
    Values closer to 100\% indicate more playable levels overall.
    A higher average indicates that more of the levels are playable.
    \item \textit{Total Nodes Explored} - This metric is the number of nodes the A* agent needed in total to reach all of the reachable gold pieces.
    We do not include nodes explored when the pathfinder attempted to reach unreachable gold pieces.
    This number can be taken as an approximation for the minimum amount of time a player would need to complete a level.
    The higher the number the longer the time needed to complete the level.  
    \item \textit{Nodes Per Gold} - Since levels do not all have the same number of gold pieces, the total nodes explored metric on it's own could potential leave an inaccurate reflection of each level.
    As such, this metric gives the average number of nodes needed to reach each reachable piece of gold per level.
    The same rule applies as with the above metric, where the larger the number the longer it potentially could take to reach each gold piece.  
\end{itemize}

\noindent
These metrics allow us to compare between our two sets of generated levels. 
We are particularly interested in the second metric as a measure of playability and the fourth metric as a measure of global coherence. 
We report the other two metrics for context to these two metrics. 
If the Snodgrass and Ontan{\'o}n levels outperform our levels in terms of playability, that could indicate that our method for placing gold pieces along the player path is flawed. 
If the Snodgrass and Ontan{\'o}n levels outperform our levels or perform similarly in terms of nodes per gold metric, this would indicate that our inclusion of the player path did not improve global coherence, and led to similarly coherent/incoherent levels as a simpler Markov chain approach.

\section{Qualitative Results}

%BIG OLD TABLE or a series of figures/boxplots

%Indicate all our results are in Table 1, and then walk the reader through the results and give our takes on those results.

%End with the support for our initial claim.

%Table 1 goes over our results. [Talk them through each column and how we're interpreting that, and make sure to include the Wilcox test showing the percent playable metrics are not significantly different]

Table 1 includes all of our results in terms of our four metrics. 
The first column gives the average and standard deviation of the total number of gold pieces included in each generated level. 
This immediately demonstrates the impact of employing a Gaussian distribution to model the total number of gold pieces in a level, as opposed to leaving it up to a Markov chain alone to place gold pieces.
The Snodgrass and Ontan{\'o}n levels have a much higher average and a much larger standard deviation.
This indicates that their levels tended to have many more gold pieces on average compared to existing Lode Runner levels, and that this number varied to a great extent with the largest number of gold pieces being nearly three times as many for the Snodgrass and Ontan{\'o}n levels.
This is an initial indication that our approach led to more coherent levels.

\begin{table*}[tbh]
\begin{tabular}{|l|c|c|c|c|}
\hline
          & Gold Total & Percentage Collected & Total Nodes Explored & Nodes per Gold  \\
          \hline
Snodgrass and Ontan{\'o}n & 18.76\rpm 15.07          & 98.94\rpm 6.98                    & 4638.93\rpm 10837.32                    & 220.62\rpm 477.89                             \\
\hline
Ours      & 7.65\rpm 2.98          & 94.68\rpm 19.07                    & 910.33\rpm 2047.28                    & 116.44\rpm 264.82             \\
\hline
\end{tabular}
\caption{Quantitative Evaluation Results}
\end{table*}

The second column of Table 1 shows the average amount of gold that can be collected per level, indicating how many of the generated levels were playable.  
These values indicate that the Snodgrass and Ontan{\'o}n levels are on average more likely to be playable. 
However, there is more complexity to this value than might first appear.
Since on average our levels have less gold than the Snodgrass and Ontan{\'o}n levels, the impact of being unable to reach a single gold piece is much higher.  
Collecting 6 of 7 available gold pieces will result in a lower percentage than 17 of 18 possible gold pieces.  
We also performed a Mann-Whitney U test to determine if these two distributions differed significantly.
The test was unable to reject the null hypothesis ($p=0.05629$) that these two distributions arose from the same underlying distribution.
Thus, we take this to mean that the difference in terms of playability was insignificant. 
As mentioned above, this runs counter to prior reported playability values \cite{snodgrass2016learning}, which is due to our choice not to model enemy locations. 
Thus, this playability metric should be considered an upper bound.

\begin{figure}[tb]
    \centering
    \includegraphics[width=3in]{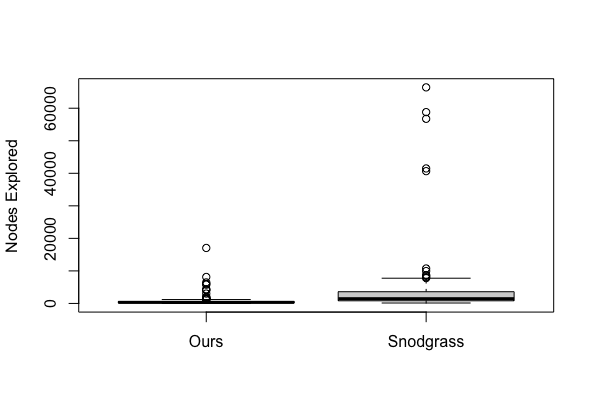}
    \caption{Boxplot Results of Nodes Explored Metric}
    \label{fig:boxplot}
\end{figure}

The third column of Table 1 gives the Total Nodes Explored metric, which we also visualize in Figure \ref{fig:boxplot} for clarity.
It is immediately clear that the Snodgrass and Ontan{\'o}n generator leads to massively more explored nodes than our approach. 
However, this is not a fair comparison due to the higher average number of gold pieces among these levels and the larger variance of the number of gold pieces. 
Thus, we turn to the fourth column and the average number of nodes explored for each gold piece. 
This comparison is closer, only indicating that the Snodgrass and Ontan{\'o}n levels require twice as many nodes to be explored to reach each gold piece. 
However, this difference is still substantial. 
This indicates that, for each gold piece, it would take a player roughly twice the time to collect it for an average Snodgrass and Ontan{\'o}n level. 
This indicates that there is no clear path between the gold pieces in the Snodgrass and Ontan{\'o}n levels.
The standard deviation also signifies that our levels are more consistent in terms of this metric.  
We take this as an indication of the greater global coherency of our generated levels, that these levels include clear paths for the player to take to reach most gold pieces, even if the placement of some of these gold pieces makes them unreachable.

Our findings suggest that both our method and the method used by Snodgrass and Ontan{\'o}n produce levels that are roughly equally playable using our A* agent that ignored enemy positions.
Looking at the metric values it would have been very difficult to improve on this metric without achieving 100 percent playability.  
Our analysis did show that we improved the consistency and reliability of the solutions to levels.  

\section{Qualitative Evaluation}

In this section, we asses the quality of our output levels in comparison to the original levels. 
We did this in order to get a more nuanced look as to whether we have actually achieved greater global coherence. 
Given the results of our first set of evaluations, it's clear that our approach was able to produce levels with clearer and more consistent solutions. 
However, it's possible that our output levels no longer resemble the original Lode Runner levels. 
For example, they may have become too simple or have lost other aspects of global coherence.
We use the following metrics to investigate this possibility:
\begin{itemize}
\item \textit{s} - The minimum size of the level to fit the path. 
Ideally, we'd like the levels to match the size of the original Lode Runner levels: a 32x22 grid.
Since we do not explicitly enforce this, our hope is that our approach will have led to an implicit bias towards levels of this size.
\item \textit{e} - The proportion of the level taken up by empty space. The original Lode Runner levels have a fair amount of variance when it comes to empty space, and empty space is often used strategically to create shapes or to indicate potential solutions. Thus, if our distribution of empty tiles matches those from the original levels, this would indicate a positive signal in terms of similar global structure.
\item \textit{i} - The proportion of the room taken up by ``interesting'' tiles that are not simply solid or empty. 
It goes without saying that the way that the Lode Runner levels employ ladders, ropes, enemies, and gold pieces is very important to the overall design of the level.
\end{itemize} 

\section{Results}

We generated 34 generated levels and compared these to the 34 original levels that we did not use to train our model.
We found that 20 of our generated levels matched the expected size of 32x22, with 14 of our levels having a smaller size.
This means that these levels could be expanded to fit the expected size while retaining their same generated structure. 
This is a positive sign, as our approach was able to implicitly lead to levels with the same or similar sizes to the original levels without explicitly modeling this constraint. 
Notably, none of the generated levels were larger than the original levels, though this may be due to the fact that we employed a constant generated path size of 103 (which was the average of the paths we extracted from the gameplay video). 
However, a generated path of 103 steps could still have led to a level larger than 32x22.

\begin{figure}[tbh]
    \centering
    \includegraphics[width = 3in]{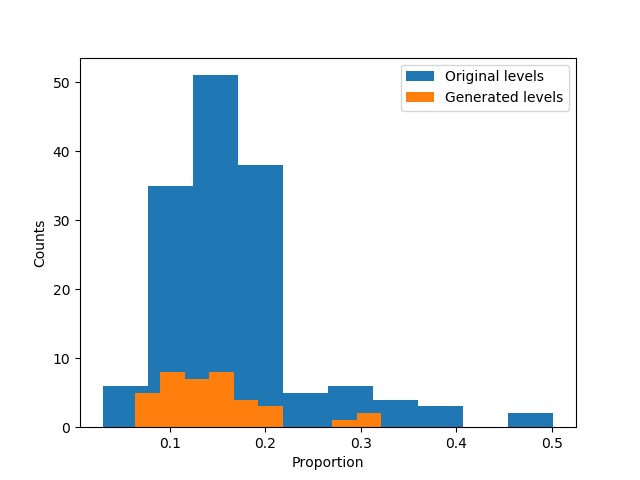}
    
    \caption{The distribution of "interesting" tiles.}
    \label{fig:figure_roi}
\end{figure}

\begin{figure}[tbh]
    \centering
    \includegraphics[width = 3in]{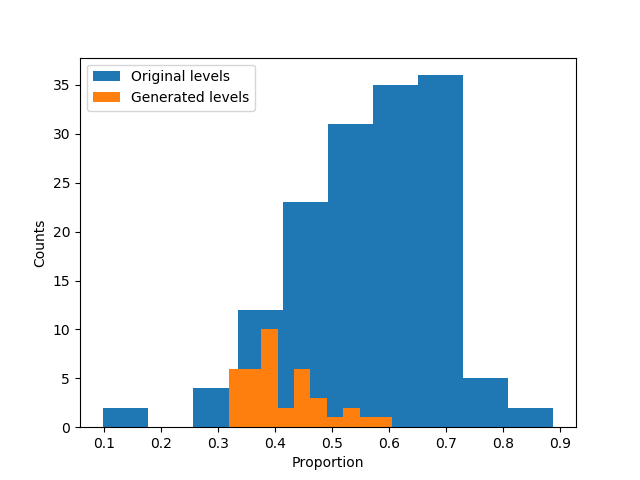}
    \caption{The distribution of empty space.}
    \label{fig:figure_space}
\end{figure}

Figure \ref{fig:figure_roi} and Figure \ref{fig:figure_space} show the distribution of ``interesting'' and empty space tiles in the generated and original levels respectively. 
The distribution of ``interesting'' tiles does suggest that our levels tended to lead to fewer interesting tiles compared to the original levels. 
However, the overall distributions are fairly similar. 
Further, it's possible that due to the 14 smaller levels the generated levels look more conservative than they truly are since this distribution does not take into account level size.
The empty space distribution seems to differ more with the original levels employing much more empty space. 
This is not an unusual problem for Markov Chain models: filling in too much content. 
However, we again note that the 14 smaller levels may be part of the problem here. 
By filling the remaining space of these 14 smaller levels with empty space, the two distributions would look much more similar.

\section{Discussion}

There are 150 levels completed in the video we used for this paper. 
However, trimming and cropping each level from the video was time-consuming and repetitive. 
Therefore we just dealt with 66 levels to extract player paths. 
If all levels were processed, the results might be improved.

Extracting path information from videos has a couple of challenges. 
Our method works well for most levels, but the levels with lots of stairs and levels with fake bricks (where there appears to be a brick, but it is actually an empty space as the player steps on it) will not work very well. 
Due to the low image quality, when the player walks passed the stairs, the combined figure becomes very difficult to recognize. 
This may sometimes lead to losing track of the player. 
The problem with the fake brick has a similar effect. 
When the player falls through a fake brick, the program fails to detect the player, hence losing track of the player.

It is a limitation of our approach that we always assume a fixed path length. 
While we made this choice for simplicity, the original levels did not all have the same path length. 
As such, it would be better to model this path length value separately. 
Alternatively, we could generate the path until we hit the desired level size and then stop.

\begin{figure}[tbh]
    \centering
    \includegraphics[width=3in]{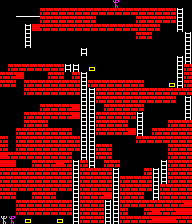}
    \caption{A bad generated level: badly placed enemies.}
    \label{fig:bad1}
\end{figure}

\begin{figure}[tbh]
    \centering
    \includegraphics[width=3in]{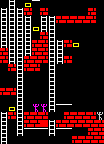}
    \caption{A bad generated level: badly designed structure.}
    \label{fig:bad2}
\end{figure}

Figure \ref{fig:bad1} and Figure \ref{fig:bad2} show two typical issues that prevent the player from completing the generated levels. 
In Figure \ref{fig:bad1}, an enemy is directly placed beside the player, and there is no other way to go around or trap the enemy. 
This shows that randomly placing enemies along the player's path is not ideal, we will also need to take consideration that the player needs to have a way to deal with enemies on the path.
We ignored this problem for our pathfinding-based evaluation, but we will need to confront it for future work.
Figure\ref{fig:bad2} had a different problem. 
The generated structure led to the player getting stuck. 
This happens when the row where the player is at and the row above it both have path information. Then when filling the tiles, the system would make a mistake that the bottom row can be bricks, and the player can dig along the upper row to create a path to the lower row.
But this will not work if the player started from the lower row.
This situation would trip up our A* pathfinder, and was one of the factors that led to our lower average playability score.

%Identify limitations of this work and what we might do to solve them

%Talk at a high level about the hope for the future of this project/what we'd like to work on and why
In this initial investigation we assumed controllability due to our input of player paths. 
While we can alter these paths and produce new levels based on them, we did not include any evaluation of this aspect of our research. 
This would be a difficult thing to evaluate without a human subject study, and so we leave it for future work.

\section{Conclusions}

%Overview of the paper, assuming the reader has read it. Basically, if you can only have someone remember 3-4 things about this paper, what should they be?
In this research project, we developed a player path-based method for the generation of Lode Runner levels. 
We extracted player paths from gameplay video to serve as training data, then used an LSTM Seq2Seq model to generate new player paths, and applied Markov Chains to produce new levels based on these paths. 
Our experimental results show that this approach can lead to improved global coherence of the generated levels while still leading to levels that share a resemblance to the original levels.
For future work, we hope to improve the proposed method to ensure playability, to use a more sophisticated pathfinding agent that takes into account things like enemy placement, and to test this approach on other games. 

\begin{acks}
This work was funded by the Canada CIFAR AI Chairs Program. We acknowledge the support of the Alberta Machine Intelligence Institute (Amii).
\end{acks}

\bibliographystyle{ACM-Reference-Format}
\bibliography{main}

\appendix

\end{document}